Open Access | Review

# Review on Monitoring, Operation and Maintenance of Smart Offshore Wind Farms

by Lei Kou [1], Yang Li [2], Fangfang Zhang [3,*], Xiaodong Gong [1], Yinghong Hu [4], Quande Yuan [5] and Wende Ke [6]

[1] Institute of Oceanographic Instrumentation, Qilu University of Technology (Shandong Academy of Sciences), Qingdao 266075, China
[2] School of Electrical Engineering, Northeast Electric Power University, Jilin City 132012, China
[3] School of Electrical Engineering and Automation, Qilu University of Technology (Shandong Academy of Sciences), Jinan 250353, China
[4] Electric Power Research Institute, State Grid Jibei Electric Power Company Limited, Beijing 100054, China
[5] School of Computer Technology and Engineering, Changchun Institute of Technology, Changchun 130012, China
[6] Department of Mechanical and Energy Engineering, Southern University of Science and Technology, Shenzhen 518055, China
* Author to whom correspondence should be addressed.

Citation Export

## Abstract

In recent years, with the development of wind energy, the number and scale of wind farms have been developing rapidly. Since offshore wind farms have the advantages of stable wind speed, being clean, renewable, non-polluting, and the non-occupation of cultivated land, they have gradually become a new trend in the wind power industry all over the world. The operation and maintenance of offshore wind power has been developing in the direction of digitization and intelligence. It is of great significance to carry out research on the monitoring, operation, and maintenance of offshore wind farms, which will be of benefit for the reduction of the operation and maintenance costs, the improvement of the power generation efficiency, improvement of the stability of offshore wind farm systems, and the building of smart offshore wind farms. This paper will mainly summarize the monitoring, operation, and maintenance of offshore wind farms, with particular focus on the following points: monitoring of "offshore wind power engineering and biological and environment", the monitoring of power equipment, and the operation and maintenance of smart offshore wind farms. Finally, the future research challenges in relation to the monitoring, operation, and maintenance of smart offshore wind farms are proposed, and the future research directions in this field are explored, especially in marine environment monitoring, weather and climate prediction, intelligent monitoring of power equipment, and digital platforms.

## 1. Introduction

Owing to concerns over the global energy crisis and air pollution, the development and utilization of wind energy, solar energy, and other renewable energy sources have been given increasingly more attention all over the world [1,2,3]. Wind energy is a form of renewable energy with mature technology that has developed rapidly in the past decades [4]. By the end of 2019, the total installed capacity of global offshore wind power reached 29.1 GW. A report on China's ability to power a huge growth in global offshore wind energy stated that the total installed capacity of global offshore wind power will reach over 234 GW by 2030 [5]. Compared with onshore wind power, offshore wind power has the advantages of high wind speed, regional climate stability, and no significant visual impact. Due to the high efficiency of offshore wind power, it is suitable for centralized development, which is an important development direction for wind power [6].

With the growing emphasis on clean energy, the installed capacity of offshore wind power has been increasing faster than ever. However, due to the particularity of the offshore wind farm environment, offshore wind farms are usually accompanied by high temperature, high humidity, high salt fog, typhoon, lightning, and so on; thus, the probability of power equipment failure is higher [7]. Meanwhile, the operation and maintenance cost of offshore wind farms is much higher than that of onshore wind farms, and the accessibility of offshore wind farms is poor [8]. Traditional operation and maintenance methods are not enough to meet the operation and maintenance requirements of smart offshore wind farms. Smart offshore wind farms need to rely on good scientific operation and maintenance strategies, intelligent fault diagnosis and monitoring technology, stable and efficient operation, and the use of maintenance ships and other advanced equipment support. Preventive operation and maintenance technologies will play an important role in the management of smart offshore wind farms and also represent the future development direction of offshore wind power operation and maintenance technologies [9]. Therefore, it is of great significance to study the monitoring, operation, and maintenance of offshore wind farms.

At present, many scholars have studied the construction, monitoring, operation, and maintenance of smart offshore wind farms [10,11]. Compared with onshore wind farms, the planning and construction requirements of offshore wind farms are relatively high. It is necessary to engage in scientific planning before construction so as to minimize their impact on the marine ecological environment. The early monitoring of safety hazards and faults of equipment in offshore wind farms is needed so as to reduce operation and maintenance costs and extend the service life of equipment. In order to reduce the operation and maintenance costs of offshore wind power, Griffith et al. [10] introduced a structural health and prognostics management system into the condition-based maintenance process with the use of a smart load management methodology; health monitoring information and economics were taken into account, but the research on relevant damage feature extraction still needed to be strengthened. Shin et al. [12] proposed an efficient methodology to design the layout of offshore wind farms in which the total cost of construction, maintenance, power loss, and other factors were considered. The inner grid layout optimizer and offshore substation location optimizer were proposed based on several optimization algorithms (k-clustering-based genetic algorithm, pattern search method, etc.), but these ignored the impact of biological factors and the geographical environment in the actual operation environment. Tao et al. [13] proposed a bi-level multi-objective optimization framework to determine the capacity of wind farms, the position of wind turbines, cable topology, etc., which consists of two inner-layer models and an outer-layer model; different aging degrees of wind turbines can be considered in the future. Du et al. [14] discussed the development process and core technology of the reliability-centered maintenance (RCM) theory and proposed an improved RCM framework for the operation and maintenance of offshore wind farms, but the impact of most environmental factors on the maintenance of offshore wind farms were ignored. Ye et al. [15] proposed a smart energy management cloud platform based on big data and cloud computing technology, and the topological structure, equipment, operation, and management of offshore wind farms were effectively integrated into the platform, which provided valuable experience in the construction and management of smart offshore wind farms, but still lacked information with regard to the expansion of the platform. Liu [16] pointed out that data communication of offshore wind farms need to rely on wireless communication techniques such as the wireless optical communication technology employed in wireless SCADA systems. However, sufficient attention must be still be given to the research on data encryption and secure transmission. Since it is difficult and time-intensive to locate short-distance transmission lines for deep-sea offshore wind farms, Wang et al. [17] proposed a Stockwell transform and random forest-based double terminal fault location method, in which the Stockwell transform method was used to extract the effective features, and random forest was used to train the data-driven classifier to classify the fault type and fault branch; however, the influence of load variation and line parameters should be further studied. Liu et al. [18] discussed some classic intelligent fault diagnosis methods for power electronic converters and proposed a random forest and transient fault feature-based fault diagnosis method for the three-phase power electronics converters, but in-depth research should also be carried out in combination with the offshore operation environment. Papatheou et al. [19] proposed artificial neural networks (ANNs) and a Gaussian process-based method to monitor the wind turbines of offshore wind farms; the proposed method was adopted to build a reference power curve for each of the wind turbines, but some additional features can be considered to improve the performance of the method. L et al. [20] proposed a Stackelberg game-based optimal scheduling modeling method for integrated demand response-enabled integrated energy systems with uncertain renewable generations, which can promote the consumption of renewable energy and reduce energy costs for users, but battery degradation and load uncertainty were ignored. In order to better realize the construction, monitoring, operation, and maintenance of offshore wind farms, more practical operation factors should be taken into account.

Around the world, governments are vigorously developing offshore wind power and have accomplished a lot in many fields. As shown in **Figure 1**, the construction and development of smart offshore wind farms mainly benefit from cloud computing, big data, Internet of Things communication, artificial intelligence (AI), and other new technologies [21,22]. This paper mainly summarizes the monitoring, operation, and maintenance of smart offshore wind farms ("offshore wind power engineering and biological and environment"), which includes environmental monitoring, power equipment monitoring, and the operation and maintenance of offshore wind farms, with some cases given.

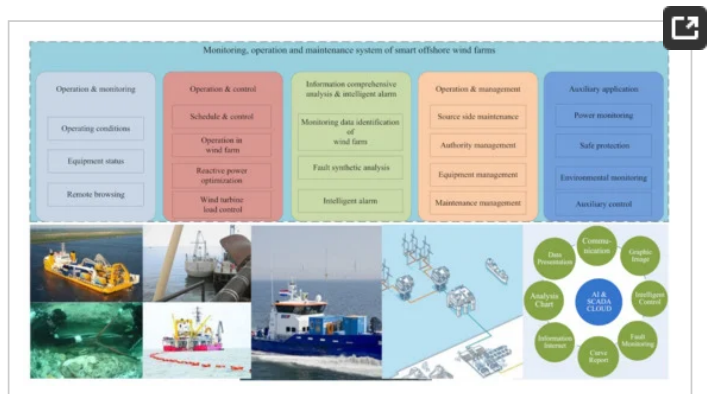

**Figure 1.** Monitoring, operation, and maintenance system of smart offshore wind farms.

The remainder of this paper is organized as follows. **Section 2** describes the environmental monitoring technologies of offshore wind farms, and some advanced equipment and technologies are also discussed. **Section 3** discusses some power equipment monitoring methods for offshore wind farms; it mainly includes the status monitoring and fault diagnosis for offshore wind turbines, power electronic converters, submarine cables, and so on. In **Section 4**, the operation and maintenance strategies of offshore wind farms are discussed in detail. Conclusions and prospects are drawn in the last section.

## 2. Environmental Monitoring for Smart Offshore Wind Farms

With the rapid development of offshore wind power, only offshore wind farms in coastal waters have had difficulty in meeting the requirements for wind energy development; these offshore wind farms have a greater impact on the marine environment [23,24]. Therefore, the study of monitoring and early warning for the marine environment, climate, natural disasters, etc., is of great significance for the healthy development of smart offshore wind farms. This section will mainly introduce some advanced marine environmental monitoring equipment and technologies in detail.

*2.1. Sea–Sky Monitoring*

Sea–sky monitoring mainly includes the climate, meteorology, floating pollutants, wind information, and some bird species, and can provide experience and optimization strategy information for the construction and operation of smart offshore wind farms in the future [25]. Sea–sky monitoring is mainly advantageous in site selection for wind farms, the planning of transmission lines, the planning of wind power generation production, the maintenance of wind turbine equipment, in considering the impact on birds, considering the safety of workers, and so on.

The machine noise, light, and magnetic field produced by offshore wind farms will have a certain impact on the foraging, breeding, and migration of birds [26,27]. For example, the offshore wind farms may directly occupy the habitat of seabirds, thus affecting their nesting and reproduction. According to [25], the research on the impact of offshore wind farms on birds mainly focuses on the behavioral, physical habitat, and direct demographic elements. According to the study in [28], the probability of a bird colliding directly with a wind turbine is very low. Fijn et al. [29] found that many birds were flying at risk height in the vicinity of the Dutch Offshore Wind farm Egmond aan Zee, but that these birds could avoid collision with the wind turbines; relevant research can also be seen in [30]. Drewitt et al. [31] studied the potential impact of wind energy developments on birds; offshore wind farms may affect the breeding, wintering, and migration of birds. The collision risk also depends on the factors related to the bird species, their number and behavior, weather conditions, and the environments of offshore wind farms (such as lighting, etc.), but the impacts of human activities should also be considered. Furness et al. [32] assessed the vulnerability of marine bird populations (especially gulls, white-tailed eagles, and northern gannets, etc.) to offshore wind farms, which found that the marine birds' long-time flight (whether they were breeding, migrating, wintering, or as prebreeders) were more likely to face the risk of collision. Niemi et al. [33] proposed an automatic bird identification system based on a fusion of radar data and image data. The data were adopted to train the classifier based on the small convolutional neural network (CNN); the classifier could then be used to monitor the bird species' behavior in the vicinity of the wind turbines, but more untrained data should be adopted to test the trained model. Gauthreaux et al. [34] proposed a fixed-beam radar and a thermal imaging camera-based method to monitor bird migration, which can be adopted to estimate the potential risk of collision between migratory birds and wind turbines, but the impact of wind turbine operation on birds should also be further considered. Plonczkier et al. [35] monitored the behavioral responses and flight changes of pink-footed geese in relation to bird detection radar so as to provide data for wind farm construction and bird protection in future, but the migration routes of other similar species still need to be studied and considered. Many scholars have put forward the use of technology for monitoring birds in order to study the birds around the offshore wind farms and give the corresponding information based on their experience for an improved construction of smart offshore wind farms and for biological protection in the future.

It is not only necessary to protect the local ecological environment, but also to monitor the local weather, wind speed, and other information in order to provide effective historical data for better operation and production in the future. Trombe et al. [36,37] performed a weather radar-based pioneer experiment to monitor the weather at the Horns Rev offshore wind farm in the North Sea, but data mining technology still needs to be considered in order to improve monitoring performance. Brusch et al. [38] analyzed severe weather by analyzing satellite images taken by space-borne radar sensors so as to provide reliable support for the operation and maintenance of offshore wind farms; more measurement data and more data fusion algorithms can be used to improve the accuracy of prediction methods in the future. Zen et al. [39] proposed an innovative use of second-level satellite products to analyze the wind speed and wave height measurements, which could help the offshore wind farm managers to make more effective strategic decisions; however, the research on the aging prediction method for offshore wind farms should also be considered. The research institute (Institute of Oceanographic Instrumentation, Shandong Academy of Sciences (IOISAS), Qingdao, China) is mainly engaged in basic research, which it has applied in the marine monitoring scientific innovation platform, the BCF handheld anemometer, scanning aerosol lidar, ship meteorological instruments, the SXZ2-2 hydrometeorological automatic observation system, underwater acoustic communication machines, and so on. **Figure 2** shows some meteorological monitoring equipment, with **Figure 2**a showing a BCF handheld anemometer that can measure wind direction, wind speed, temperature, humidity, orientation, atmospheric pressure, and GPS coordinates at the same time [40]. **Figure 2**b shows the scanning aerosol lidar, which can realize the observation of dust, haze, rainfall, and other types of weather. **Figure 2**c shows the ship meteorological instrument, which can measure and display meteorological parameters such as wind speed, wind direction, air temperature, relative humidity, air pressure, visibility, and cloud bottom height in real time. **Figure 2**d shows the SXZ2-2 hydrometeorological automatic observation system, which can be installed on various marine stations and offshore observation platforms, and can realize the automatic observation of tide, wave, surface temperature, salt, air pressure, temperature, relative humidity, precipitation, visibility, water quality, and other parameters. **Figure 3** shows the long-term observation system of air-sea coupling in Greenland, which can obtain air-sea coupling data, improve the long-term prediction level of ocean and climate, and improve the accuracy of climate prediction [40]. In addition, the establishment of a marine meteorological characteristics data acquisition station in offshore wind farms is very important; the wind anemometer, wind vane, and other related marine equipment are used to collect marine meteorological data so as to more effectively guide the operation and maintenance of smart offshore wind farms, wind turbine group work safety level assessment, and other marine operations in the future.

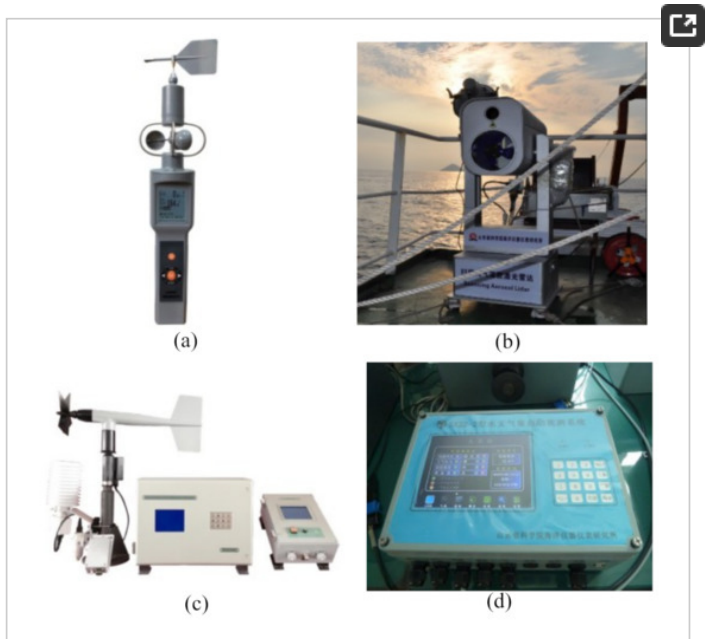

**Figure 2.** Some meteorological monitoring equipment: (**a**) BCF handheld anemometer; (**b**) Scanning aerosol lidar; (**c**) Ship meteorological instrument; (**d**) SXZ2-2 Hydrometeorological automatic observation system.

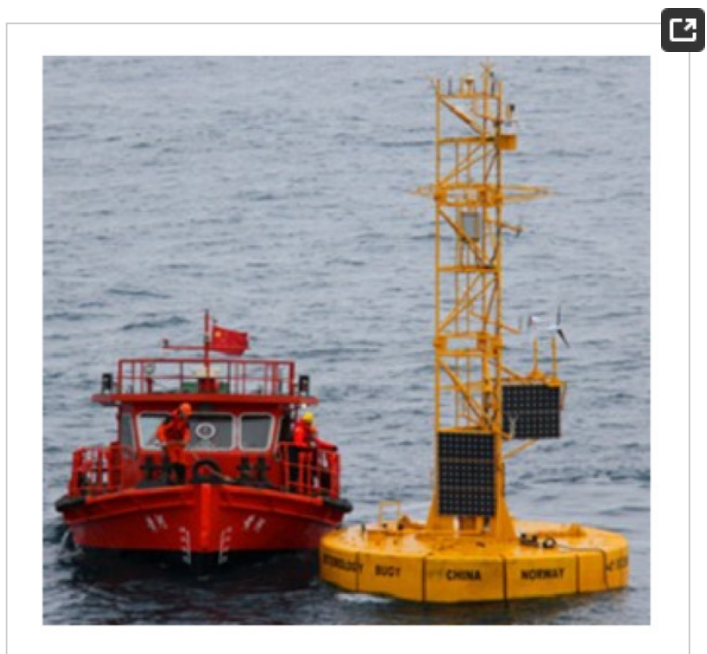

**Figure 3.** Long-term observation system of air-sea coupling in Greenland.

*2.2. Sea Surface Monitoring*

Monitoring of the sea surface mainly includes the measurement and evaluation of offshore wind energy resources, marine ecological protection and construction planning, global marine environmental protection, maritime search and rescue, emergency monitoring for red tide and sea ice, and other measures in disaster prevention [41,42,43,44,45,46]. It is of great importance to optimize the production scheduling, operation, and maintenance strategy of offshore wind farms and to protect the safety of workers.

Wind energy resource is an important factor affecting the economy of offshore wind farms, and the measurement and evaluation of wind energy resources is the key to the success of wind farm construction [47,48]. Sea surface roughness is an important parameter affecting the evaluation of offshore wind energy [49,50]. Different from land roughness, sea surface roughness is unstable, which mainly depends on the size of real-time waves [51,52,53]. The interaction between wind and waves is affected by water depth, wind speed, offshore distance, and other factors [54]. **Figure 4** shows the SBF series coastal telemetering wave gauge, which can realize automatic wave measurement in coastal stations, ports, islands, offshore platforms and ships, among others [40]. Lin et al. [54] proposed a new parameterization based on observations to estimate sea surface roughness variations according to wind speed and sea state, but there are many other factors that should be considered (such as other parameterizations for the drag coefficient). Bao et al. [55] introduced the multi-incidence maximum likelihood estimation method to the inversion of sea surface wind speed by precipitation radar, whose error is very close to that of the buoy, while the AI-based methods can be further considered for wind speed prediction. Li et al. [56] proposed a surface current inversion method based on the high-frequency distributed hybrid sky–surface wave radar, in which the unknown ionospheric state was regarded as a black box, and the key parameters are extracted to calculate the surface current on the basis of the scattering model; however, the real-time ionospheric model still needs to be considered. Wu et al. [57] studied the relationship between sea surface wind speed changes and sea surface temperature in the South China Sea region during the passage of typhoons from May to October in 2000–2010; the Atmospheric profiles should be taken into account in the future. Li et al. [58] proposed a new Geophysical Model Function XMOD2, which can deduce the sea surface wind speed based on the TerraSAR-X data, but the comparison between the scatterometer and microwave radio measurements needs to be further studied. Ebuchi et al. [59] evaluated the all-weather sea surface wind speed product with airborne Stepped Frequency Microwave Radiometers data, but the effect of negative bias needs to be further eliminated. Bi et al. [60] proposed a method based on feature-selective validation to extract and evaluate one-dimensional dynamic sea surface features, in which the Monte Carlo method was employed to establish the dynamic sea surface model, and the relationship between sea surface height fluctuation and different wind speed was simulated and analyzed; rough sea surface electromagnetic scattering can be studied in the future. Tauro et al. [61] proposed a microwave radiometer's (MWR) sea surface wind speed retrieval algorithm, which can use the numerical weather prediction estimation of wind direction to correct the MWR surface brightness temperatures; nevertheless, the standard deviation of the retrieved wind speed can be further eliminated. Galas et al. [62] introduced some GNSS-based precise technologies in which the GNSS-equipped surface buoys could be applied to monitor the sea surface roughness and sea level, but the accurate reflection analysis of ocean altimetry is limited by ocean roughness; an accurate observation of ocean roughness can be considered to solve this problem. Hou et al. [63] adopted a marine buoy that was placed within the radar coverage to monitor the sea states (wind speed, surface current, etc.), but the model accuracy still needs further verification in more complex sea conditions, and an even longer-term field observation is required. Zhou et al. [64] found that sea surface wind speeds (SSWS) are usually related to wind-induced oriented textures and proposed an SSWS retrieval model to retrieve sea surface wind directions, but a more complete hurricane model should be used for in-depth research so as to improve the performance of the method. Ren et al. [65] proposed an empirical Ku-band low incidence model-2(KuLMOD2), which can be used to retrieve and verify sea surface wind speeds form the interferometric imaging radar altimeter (InIRA) data; the retrieval errors can be further eliminated, and the validation data are also limited. Through research and the monitoring of sea surface roughness, the locations of smart offshore wind farms can be better selected. However, we should also strengthen the monitoring of complex marine environments and improve the monitoring accuracy.

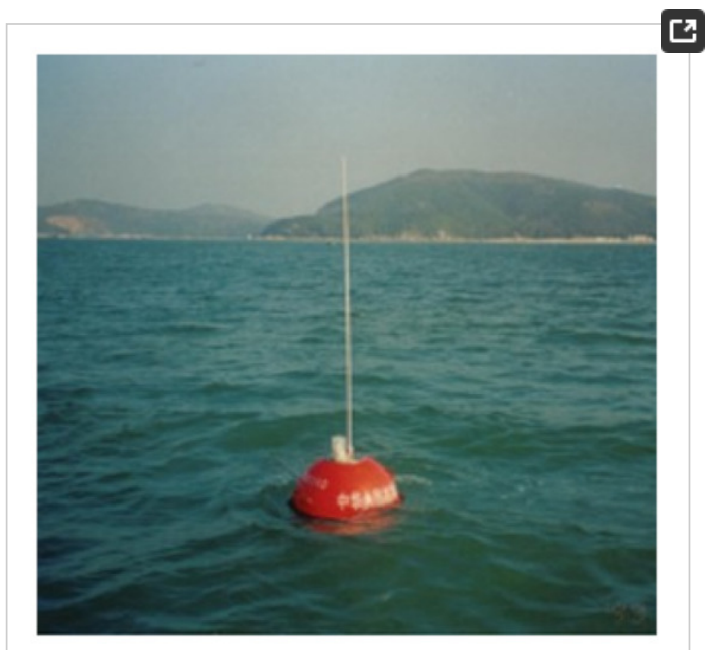

**Figure 4.** SBF series coastal telemetering wave gauge.

The monitoring of marine natural disasters and environmental pollution is of great significance to the construction of smart offshore wind farms, especially of storm surges, red tide, oil spills, sea ice, and so on [66,67,68,69]. **Figure 5** shows images of natural disasters and environmental pollution [40]. **Figure 5**a shows how a sea ice disaster affects human activities and the safe operation of facilities on the coast and sea, especially events that cause the loss of life, resources, and property such as channel blocking, marine facilities and coastal engineering damage, harbor and wharf freezing, aquaculture damage, etc. Sea ice monitoring is very important for vessel navigation, equipment maintenance planning, and weather forecasting in smart offshore wind farms. Shen et al. [70] studied and evaluated the sea ice detection method based on some machine learning methods and selected the more suitable features and algorithms; in addition, feature engineering should be deeply studied to improve the accuracy and adaptability of classification methods. Gelis et al. [71] proposed a Fully Convolutional Network-based method to monitor the sea ice concentration; it could generate sea ice concentration maps from Sentinel-1 Synthetic Aperture Radar (SAR) images, but more validation data sets in different situations need to be used to validate the method so as to ensure the effectiveness of the method. Ren et al. [72] proposed a deep learning model-based method to classify the sea ice and open water from SAR images. The SAR images were employed to train the deep learning model, but more SAR images should be collected to evaluate the model. Song et al. [73] proposed a combined learning of temporal and spatial features, residual CNN, and long short-term memory (LSTM) network-based method to classify the SAR images of sea ice; however, the data of coastal land should be considered to improve the adaptability of the model, and the model parameters can also be optimized. **Figure 5**b shows an oil spill in the process of offshore wind farm construction; the foundation of the wind turbine is driven directly into the sea floor. The laying of the submarine power transmission cable also requires deep trench excavation, which can lead to suspended sediment on the sea floor; meanwhile, some sediment may be agitated, causing the water to be turbid. Consequently, the water quality of the sea area will be polluted due to the careless spill of some oily wastewater. Ren et al. [74] proposed a one dot fuzzy initialization strategy to detect marine oil spill regions, which did not need to label multiple pixels to initialize energy minimization. The method can be used to process SAR polarimetric feature maps in the future so as to detect oil leakage more effectively. Singha et al. [75] developed an offshore monitoring platform in which the extracted features from SAR images were used to train the support vector machine-based (SVM) classifier in order to detect the oil spills; nevertheless, the method of removing redundant features should be considered to be able to select more effective features so as to improve the computational performance. Mdakane et al. [76] developed a monitoring system based on a gradient-boosting decision tree (GBT) classifier in which multiple oil spill features were used to train the GBT classifier to automatically detect oil spills, but the impact of instrument-dependent and spatial resolution-dependent parameters still need to be further studied. Garcia-Pineda et al. [77] proposed a Textural Classifier Neural Network algorithm (TCNNA) to detect oil spills; here, the SAR data and wind model outputs were each processed by two neural networks. Lee et al. [78] proposed a recursive neural network-based method that can eliminate the pixels corresponding to the ship and ship shadows in the satellite images and subsequently detect the oil spill. However, more external environmental factors should be considered to improve the adaptability of the method in [77,78]. **Figure 5**c shows a storm surge; storm surge disasters are usually caused by typhoons, extratropical cyclones, cold fronts, sudden change in air pressure, and so on, which can easily cause the loss of life and property. Storm surge monitoring will allow for the better planning of operation and maintenance strategies as well as protect the lives of the workers. Geng et al. [79] adopted 2-h GPS positions at 26 stations around the southern North Sea to identify the loading displacements caused by the storm surge. Wang et al. [80] proposed a deep reinforcement learning-based storm surge flood simulation method, which provides reliable data for preventing storm disasters, but more actual data are needed and should be used to verify the effectiveness of the method. **Figure 5**d shows the red tide; the main harm inflicted by the red tide is the destruction it causes in the marine environment, the death of many marine and mariculture organisms, and the damage created in fisheries and aquaculture. It may cause huge economic losses and seriously affect people's lives. Huang et al. [81] established a loop-mediated isothermal amplification (LAMP) and lateral flow dipstick (LFD) method, which can quickly detect the Karenia mikimotoi (a common nearshore red tide alga). Qin et al. [82] proposed a red tide time series forecasting method on the basis of the Autoregressive Integrated Moving Average (ARIMA) and the deep belief network. More actual complex operation scenario data should also be used to improve the effectiveness of the method in [81,82]. In addition to monitoring the natural disasters and environmental pollution, many scholars have studied the methods of maritime search and rescue and have had some achievements [83,84,85,86]. For example, Yang et al. [83] proposed a search and rescue solution based on exploration path planning and ad hoc group networking methods, in which unmanned aerial vehicles and unmanned surface vehicles were adopted in co-operative search and rescue activities. Through sea surface monitoring, a better operation and maintenance plan can be made, which can thus reduce economic losses and protect the lives of workers.

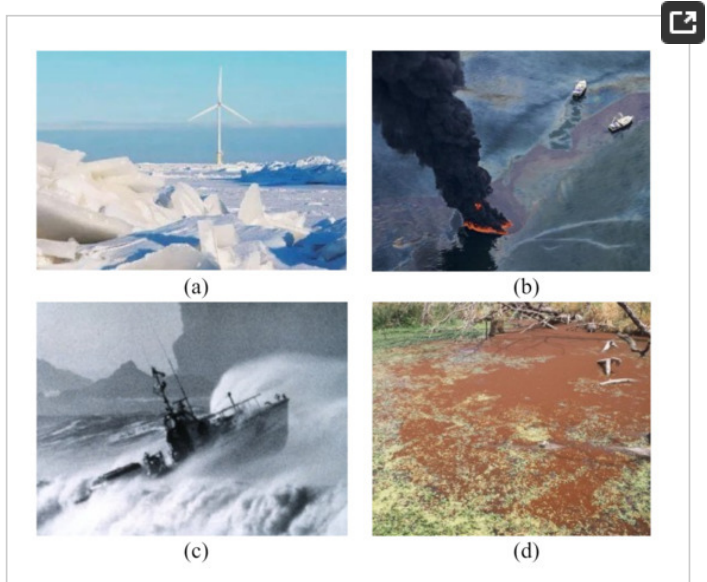

**Figure 5.** Natural disasters and environmental pollution: (**a**) sea ice; (**b**) oil spill; (**c**) storm surge; (**d**) red tide.

*2.3. Sea Floor Monitoring*

The offshore wind farms may damage the sea floor environment and cause the death of benthos. This section will discuss the advanced monitoring equipment for the sea floor environment, earthquake monitoring, benthos monitoring, large marine organism monitoring (dolphins, etc.), and other advanced technologies [87,88,89,90].

Some researchers have found that offshore wind turbines do cause some damage to marine organisms [91,92]. For example, (1) the sound of piling during the construction of wind turbine infrastructure may cause damage to the hearing of marine animals; (2) the noise of the wind turbine may affect the communication or sense of direction of marine animals or fish, causing them to get lost; (3) in the process of offshore wind power construction and maintenance, the operation of vessels may also interfere with the habitat of marine fish. **Figure 6** shows the underwater acoustic modem (IOISAS Seatrix), which can be used in underwater communication, earthquake monitoring, biological monitoring, and other fields [40,93,94].

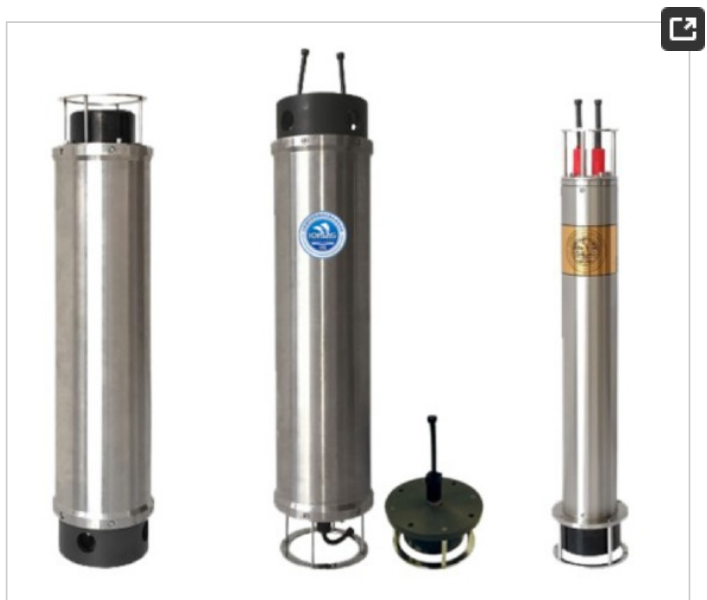

**Figure 6.** IOISAS Seatrix.

**Figure 7** shows the underwater junction box observation network system, which mainly includes the docking of underwater vehicles, data communication relay, underwater data acquisition, control command transmission, etc. [40]. The application fields of the system include marine environmental monitoring, marine disaster prediction, marine geological mapping, marine resource exploration, and so on [95]. Huang et al. [96] designed a pressure self-adaptive water-tight junction box (PSAWJB) in which a redundancy design method was employed to improve its reliability. Huang et al. [97] proposed a pre-compression method to improve the pressure compensation performance of the film-type pressure self-adaptive watertight junction box. More activities should also be carried out in the marine environment so as to improve the designed instruments in [96,97].

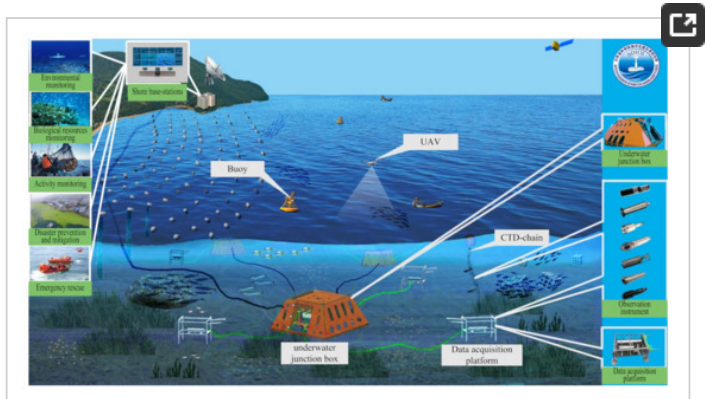

**Figure 7.** Underwater junction box observation network system.

Although the harsh marine environment brings a lot of inconvenience to the operation and maintenance of smart offshore wind farms, the application of underwater robots and unmanned aerial vehicles (UAVs) improves the convenience of operation and maintenance activities as well as reduces the safety risk for workers. Therefore, a robotic system is also a key part of smart offshore wind farms. The underwater environment is dangerous and complex, and robots can stay in the water for a longer time or work in a deeper environment as compared with human beings [98]. **Figure 8** shows some robots with different functions, which are supported by Alphaer (Shenzhen, China) Technology Co., Ltd. **Figure 8**a shows a spraying robot, which can replace the manual delivery of goods, target testing, monitoring, operation, processing, and so on. **Figure 8**b shows a small diameter pipe robot, which can carry relevant equipment and sensors to detect or clean the environment inside of the cable ducts. **Figure 8**c shows an underwater vehicle ROV II, which can be used to explore the underwater environment, check on resources, hydrology, fishery as well as to investigate the underwater coral reef ecology and other underwater operations. **Figure 8**d shows a ROS robot, which can build a map and detect a specific environment in the room, and can complete the regular inspection task. Xu et al. [99] developed an uncalibrated visual servoing scheme, which can be used for the precise positioning of underwater soft robots. Debruyn et al. [100] proposed robust technology for a multirotor and underwater micro-vehicle-based method, which can be used for automated water sampling in difficult-to-reach locations. Cai et al. [101] proposed a sphere cross section-based 3-D obstacle avoidance algorithm, which can be used for an autonomous underwater robot. However, the problem of communication between multiple underwater robots still needs to be further studied in [99,100,101]. Thus, the automated monitoring of the sea floor is an ideal means of protecting the marine ecological environment as well as the workers' lives.

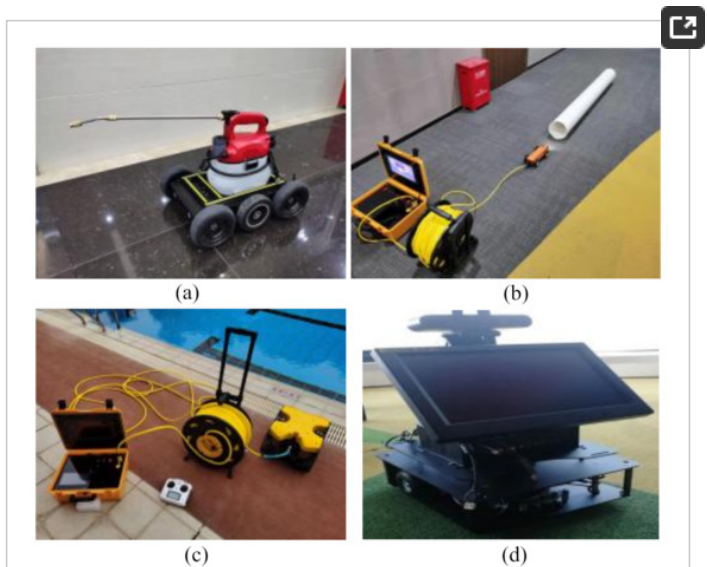

**Figure 8.** Some robots with different functions: (**a**) Spraying robot; (**b**) Small diameter pipe robot; (**c**) ROV II; (**d**) ROS robot.

## 3. Power Equipment Monitoring for Smart Offshore Wind Farms

Since the offshore wind farm environment is harsh and complex, the equipment fault rate of offshore wind farms is significantly higher than that of onshore wind farms [102,103,104]. Therefore, strengthening the research on monitoring and fault diagnosis for offshore wind farm equipment can improve the utilization rate of equipment, prolong the service life of equipment, reduce down time, increase the operation safety, and greatly improve the competitiveness of offshore wind power [105,106]. Monitoring and fault diagnosis for offshore wind turbines, power electronics converters, submarine cables, and other equipment will be discussed in detail in this section.

### 3.1. Monitoring for Offshore Wind Turbines

The structure of an offshore wind turbine is basically the same as that of an onshore wind turbine, which is mainly composed of a foundation, tower, nacelle, hub, wind wheel, drive train system, gearbox, generator, brake system, pitch system, yaw system, sensors system, electrical system, control system, communication system, and so on [107]. As shown in **Figure 9**, the common faults are mainly concentrated in several key components such as the gearbox, generator, tower, blades, and foundation. Once any of the components has a functional fault, the wind turbines may shut down, which will affect power generation and cause economic losses. Therefore, it is necessary to carry out the condition monitoring and fault diagnosis for offshore wind turbines to reduce the fault rate and maintenance cost, and to ensure the safe and efficient operation of offshore wind turbines [108,109,110].

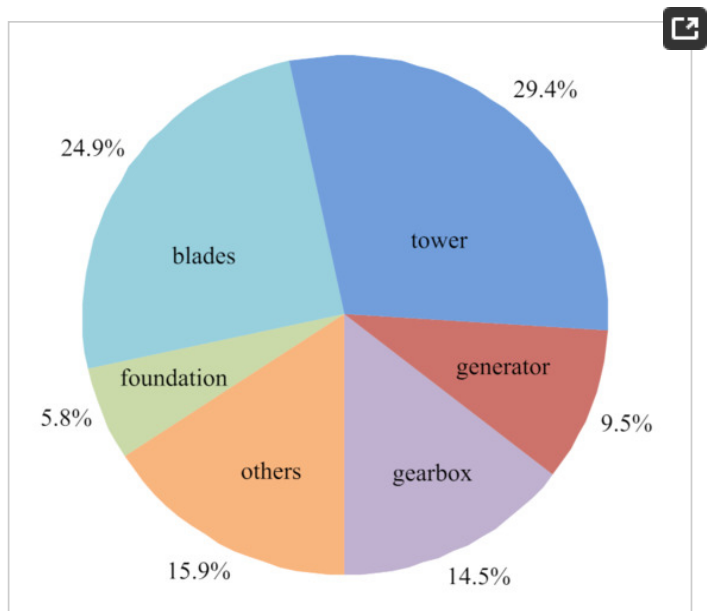

**Figure 9.** Downtime distribution of each part.

The monitoring data source is composed of all kinds of sensors installed on the equipment, and the main signals monitored are the vibration, acoustic emission, generator speed, stress, torque, temperature, oil, electrical signal, SCADA data, and so on [111,112,113].

The gearbox is an important part which often causes the downtime of wind turbines, and the fault diagnosis of gearboxes has been a concern of many scholars. Cheng et al. [114] proposed a deep learning based fault diagnosis method for wind turbine drivetrain gearboxes, in which a stacked autoencoder and a support vector machine were used to train the fault classification; the fault diagnosis flowchart is as shown in **Figure 10**. Cheng et al. [115] proposed a fault diagnosis method based on a doubly fed induction generator (DFIG) stator current envelope analysis for wind turbine drivetrain gearboxes, in which the synchronous resampling algorithm was the Hilbert transform; power spectral density analysis was used to extract fault features. Yu et al. [116] proposed a fault diagnosis method based on a fast deep graph convolutional network for wind turbine gearboxes, in which the original vibration signals were decomposed by a wavelet packet, and graph convolutional networks were used to extract the features. In [114,115,116], it is also necessary to consider more information such as operating conditions and equipment parameters in order to ensure the effectiveness of the method. Cheng et al. [117] proposed an adaptive neuro-fuzzy inference system (ANFIS) and particle filtering (PF)-based fault prognostic and remaining useful life (RUL) prediction method, in which the ANFIS was adopted to extract fault features, and the PF algorithm was used to predict the RUL of the gearbox; the noise-to-signal ratio features can be considered to improve the performance of the method in future. Yang et al. [118] proposed a deep joint variational autoencoder (JVAE)-based method to detect gearbox faults, in which the wind farm supervisory control and SCADA data were used to train the data-driven classifier, but the JVAE network architecture needs to be further improved to enhance the performance of fault diagnosis. Jiang et al. [119] proposed a multiscale convolutional neural network (MSCNN)-based fault diagnosis method for a wind turbine gearbox, in which the vibration signals were used to train the MSCNN classification model. Jiang et al. [120] proposed a feature representation learning method (stacked multilevel denoising autoencoders), which can be used to extract features and classify them according to the complex vibration signals of wind turbine gearboxes. In [119,120], the data sets under different operating conditions and the problems of imbalanced data distribution can be further studied in the future so as to ensure the practicability of the algorithm. Yoon et al. [121] proposed a piezoelectric strain sensor-based fault diagnosis method for planetary gearboxes, which has been validated on sun gear, planetary gear, ring gear, and so on; however, the effects of electrical faults should also be considered in subsequent studies. Du et al. [122] proposed a fault diagnosis method on the basis of the union of redundant dictionary for wind turbine gearboxes, in which an adaptive feature identification method was used to extract multiple components from the superimposed signals. Pu et al. [123] proposed a deep enhanced fusion network (DEFN)-based fault diagnosis method for wind turbine gearboxes, in which the fused three-axis features were used to train the DEFN model. In [122,123], the scalability and generality of the algorithm should be considered in future. Lu et al. [124] proposed a current-based fault diagnosis method for drivetrain gearboxes, in which a statistical analysis algorithm was used to extract the fault features from the nonstationary stator current signals; nevertheless, the fault type identification, different fault locations, and the remaining useful life prediction should also be considered. He et al. [125] proposed an unsupervised feature learning-based fault diagnosis method for gearboxes; meanwhile, a multiview sparse filtering (MVSF) method was adopted to extract current features. Fault feature extraction under non-stationary conditions still needs to be studied so as to improve the practicability of the diagnosis methods. Through the monitoring and fault diagnosis of gearboxes, maintenance for gearboxes can be carried out in time to avoid downtime and huge economic losses.

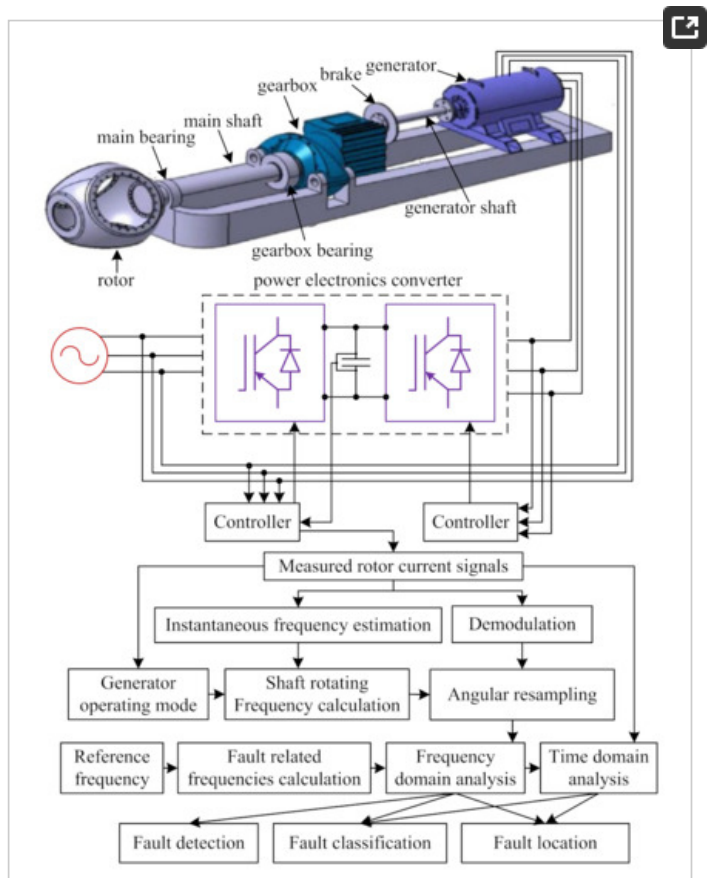

**Figure 10.** Gearbox fault diagnosis flowchart.

Generator fault is one of the main factors that lead to the wind turbine shutdown, which is why generator fault diagnosis has been a hot research topic [126,127]. The early detection of generator fault is very important for the complex system, which can save time and cost and also help to take the necessary measures to avoid dangerous situations [128,129]. Because of the lack of early warning time and fault samples of the offshore SCADA system, Wei et al. [130] proposed a stacking fusion algorithm framework for the early warning and diagnosis of offshore DFIG (as shown in **Figure 11**); a fault-tolerant operation is worthy of further research. Zhang et al. [131] proposed a SCADA data-driven method in which the subspace reconstruction-based robust kernel principal component analysis (SR-RKPCA)-based method was used to extract nonlinear features from the SCADA data. Wang et al. [132] proposed a multiscale filtering spectrum based fault diagnosis method in which the current and vibration signals were used in the diagnosis of bearing fault of direct-drive wind turbines. Jin et al. [133] proposed an ensemble fault diagnosis method for wind turbine generators in which the ensemble method was adopted to analyze the SCADA time series data. In [131,132,133], the influence of equipment parameters on fault features should be considered in the future. Watson et al. [134] proposed a condition monitoring method for DFIG in which the wavelet was used to extract fault features, but the study should also consider the impact of different operating environments and different equipment on the samples. Gong et al. [135] proposed a current-based mechanical fault diagnosis method in which an impulse detection algorithm was adopted to detect the faults, but the actual operation data should be considered to improve the method so as to improve its practical application value. Wang et al. [136] proposed a time-varying cosine-packet dictionary-based fault diagnosis method for wind turbine bearings in which the shaft rotating frequency was used to extract fault features form the vibration signals; the domain knowledge can be considered to extract more adaptive fault features to improve the effectiveness of diagnosis methods in the future. Gong et al. [137] analyzed generator stator fault currents and proposed a current-based bearing fault diagnosis method in which only a one-phase stator current signal was used. Wang et al. [138] proposed a current-aided vibration order tracking-based bearing fault diagnosis method in which the reference signal was extracted from the stator current signal. In [137,138], more fault problems in actual complex operation conditions should be considered. Jin et al. [139] proposed a generator current signal and correlation dimension analysis-based quantitative health condition evaluation method in which the fault features were extracted from the current signals, but the scalability of the method to different types of wind turbines should be considered. Wang et al. [140] proposed a PCA and ANN-based condition monitoring method that can locate the faults of wind turbines (the gearbox fault and the generator-related fault); a real-time online monitoring method should be considered in the future. With large-scale wind turbines put into operation, the number of generator faults increases. In order to ensure the safe and efficient operation of smart offshore wind farms, it is of great significance to conduct further research on state monitoring and fault diagnosis for generators.

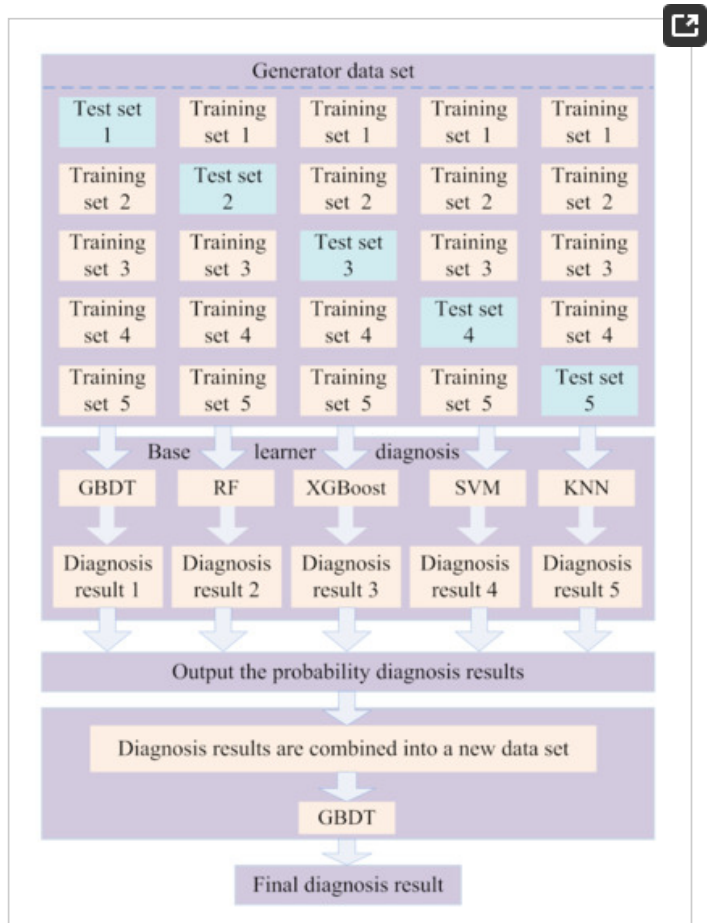

**Figure 11.** Stacking fusion algorithm framework (RF: Random Forest; SVM: Support Vector Machines; KNN: K Near Neighbor; GBDT: Gradient Boosting Decision Tree).

In addition, some scholars have studied condition monitoring and fault diagnosis for towers, blades, foundation, sensors, and so on [141,142,143,144,145]. Since the tower bears the harsh wave and wind loading conditions for a long time, Li et al. [146] proposed an inverse finite element-based structural health monitoring method for offshore wind turbine towers. Liu et al. [147] proposed an iterative nonlinear filter based fault diagnosis method for wind turbine blade bearings. In [146,147], future research can focus on other components of offshore wind turbines to realize a complete and practical monitoring system. In order to improve the stability of the wind turbine system, Peng et al. [148] proposed a wireless sensor network based fault diagnosis method for sensor faults, short faults, noise faults, and so on; however, research on wireless data security encryption should be strengthened in the future, and advanced encryption technologies such as chaotic encryption can be considered to ensure data security. Several fault diagnosis methods are conducive to the improvement of the overall stability of offshore wind turbines and reduce the costs of operation and maintenance.

*3.2. Monitoring for Power Electronic Converters*

With the development of large-scale offshore wind power, AC transmission technology will be limited by the transmission distance. DC transmission technology will become the development direction of offshore wind power long-distance transmission, especially the flexible high-voltage direct current transmission, which can automatically adjust the voltage, frequency, power, and so on [149]. For example, DC transmission technology has been used in the BorWin1 offshore wind farm in Germany and the Nan'ao VSC-MTDC Project in China [150,151]. With the wide application of power electronic converters, the problem of fault diagnosis has become more and more prominent. Therefore, it is of practical and economic significance to study the monitoring and fault diagnosis technology of power electronic converters, which can avoid the occurrence of secondary faults and reduce maintenance time [152,153].

Although there are various means to improve the reliability of the power electronic converter system, the fault is still difficult to avoid [154,155]. In 2007, the fault rate or outage rate caused by the electrical system (converters, control system, etc.) was high at the Egmond aan Zee offshore wind farm in the Netherlands, faults of power semiconductor devices, which mainly include short-circuit faults and open-circuit faults [157]. Since a short-circuit fault is very destructive, it is difficult to realize the IGBT short-circuit fault diagnosis and protection based on the software algorithm, and the short-circuit faults are protected by the standard hardware circuit; IGBT open-circuit faults will not cause serious over-current or over-voltage in a short time, can last for a period of time, and will not trigger the hardware protection system [158,159].

The short-circuit fault is mainly caused by overheating, over-voltage breakdown, wrong driving signal, etc. Moreover, it is destructive and easy to burn other components of power electronic devices. The hardware protection methods for IGBT short-circuit faults mainly include the desaturation detection method [160], inductance detection method [161], collector current detection [162], etc. Since a fast fuse has the characteristics of small heat capacity, it can be fused before the fault current reaches the preset short-circuit current. In order to reduce the harm of a short-circuit fault, Abdelghani et al. [163] used two fast fuses to convert the short-circuit fault into an open-circuit fault (as shown in **Figure 12**). In this case, it is more significant to improve the diagnosis of open-circuit faults of power electronic converters.

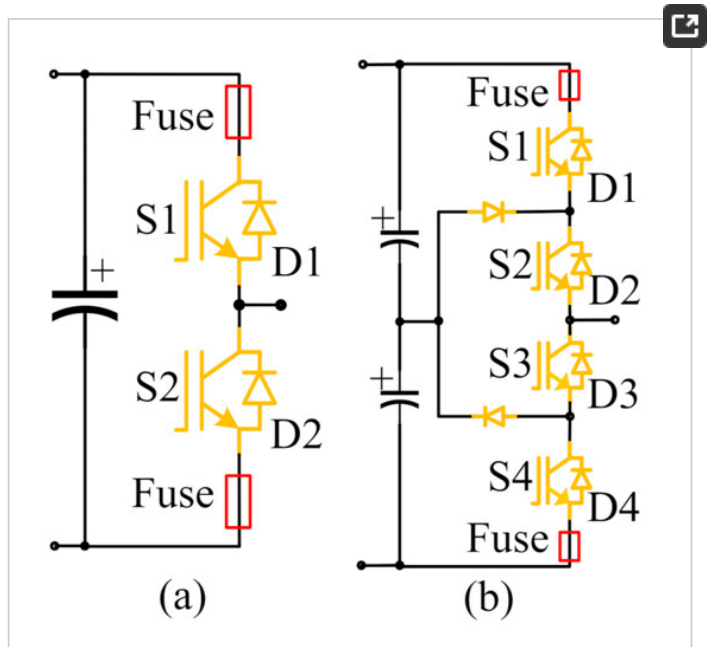

**Figure 12.** Short-circuit fault isolation technology with fast fuses: (**a**) Two-level; (**b**) NPC.

Generally, the main causes of IGBT open-circuit faults are device fracture, binding wire fracture or welding off, poor wiring, circuit faults, etc. [164]. According to [165], when the open-circuit faults happen in IGBTs, the bypass diode can still work normally, and the power electronics converters will not shut down immediately, which will lead to the increase of current and voltage harmonic content and reduce the power supply quality. However, the IGBT open-circuit fault may not be found for a long time, resulting in secondary damage or catastrophic faults of other equipment. Power electronic converters are mainly composed of power semiconductor devices, and the systems are not linear, which limit the application of an open-circuit fault diagnosis method based on a fault mathematical model [166]. The data-driven fault diagnosis method does not need to establish an accurate mathematical model of power electronic converters, where the typical methods include: ANN, time series prediction, SVM, random forests (RFs), PCA, or other AI-based fault diagnosis methods. AI technology has the self-adaptive learning ability from fault samples, which can realize the mapping between fault data and fault state and obtain the mature fault diagnosis classifier (as shown in **Figure 13**). Then, the mature fault diagnosis classifier can locate the faults in power electronic converters.

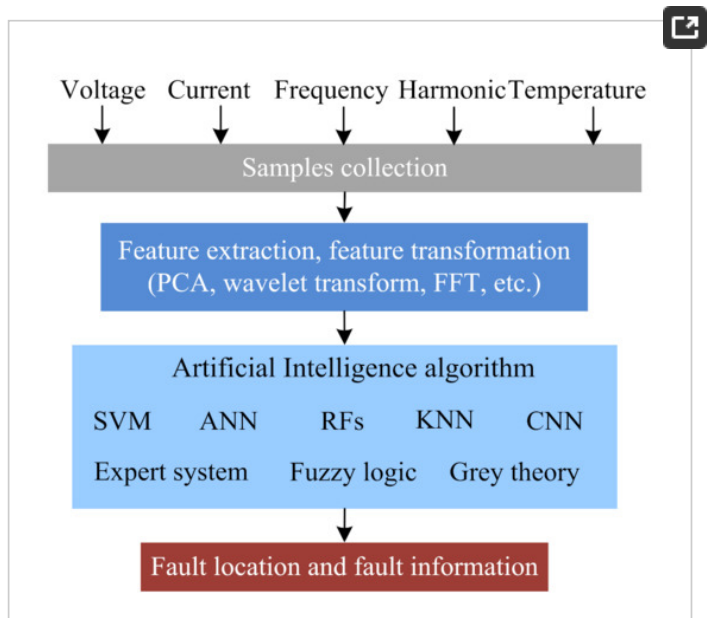

**Figure 13.** AI(Artificial Intelligence)-based open-circuit fault diagnosis methods(ANN: Artificial Neural Network; CNN: Convolutional Neural Networks).

With the development of the smart grid, the data-driven fault diagnosis technology of power electronic converters has become a research hotspot in the industry [167,168,169]. Wang et al. [166] proposed a knowledge data-based fault diagnosis method for three-phase power electronic energy conversion systems in which the knowledge-based method was used to extract the fault features, and the data-driven method was used to train the fault diagnosis classifier; the fault diagnosis schematic is as shown in **Figure 14**. Xia et al. [167] proposed a data-driven fault diagnosis method for three-phase PWM converters in which the three-phase AC current signals, FFT, and ReliefF algorithm were adopted to extract features, and a sliding-window classification framework was used to improve the diagnosis performance. In [166,167], the influence of diode faults and sensor faults can be considered in future research. Cai et al. [168] proposed a Bayesian network-based fault diagnosis method for three-phase inverters in which the FFT was used to extract the signal features from the output line-to-line voltages; a wavelet transform can be considered to realize the signal feature extraction in the future. Li et al. [170] proposed a model data hybrid-driven fault diagnosis method for power converters in which the model information and ANN were combined with the diagnosis robustness and diagnosis speed, but the effectiveness of the method should also be verified and adjusted through different complex topology applications. Xue et al. [171] proposed a multilayer LSTM network-based fault diagnosis method for back-to-back converters in which three-phase currents and voltage signals were used to train the data-driven fault diagnosis classifier; the LSTM network can be continuously improved to adapt to different systems and new complex fault scenarios in the future. Kiranyaz et al. [172] proposed a one-dimensional CNN-based fault detection and identification method for modular multilevel converters (MMC) in which the raw voltage and current data were used to train the CNN classifier; the method can also be implemented and verified in larger and more complex topology and validated in real-time performance in the future. Li et al. [173] proposed a mixed kernel support tensor machine (MKSTM) fault diagnosis method for MMC in which the AC current and internal circulation current were used to classify the fault locations, but the method ignores many nonlinear noises in the actual system; it should be further verified in the actual operation system. Huang et al. [174] proposed a data-driven fault diagnosis method for photovoltaic inverters in which the multistate data processing block was used to distinguish different features, the subsection fluctuation analysis block was adopted to extract fault features, and ANN was used to realize intelligent classification. Khomfoi et al. [175] proposed an AI-based fault diagnosis and reconfiguration method in which the PCA, genetic algorithm, and neural network were used to implement the fault diagnosis classifier for a cascaded H-bridge multilevel inverter. In [174,175], the influence of load faults and diode faults on the fault features should be considered in future research so as to make the method more practical. Kame et al. [176] proposed an adaptive fault diagnosis method for a single-phase inverter based on a neuro-fuzzy inference system algorithm in which the inverter output current was used as the monitoring signal to locate the faults. Stonier et al. [177] proposed an ANN based controller to diagnose the open-circuit faults of a solar photovoltaic (PV) inverter. In [176,177], the grid-connected system was considered in their methods, and the influence of other system faults on the fault features should be considered in the future. Monitoring and fault diagnosis technology can avoid secondary faults or catastrophic faults, which is of great significance to

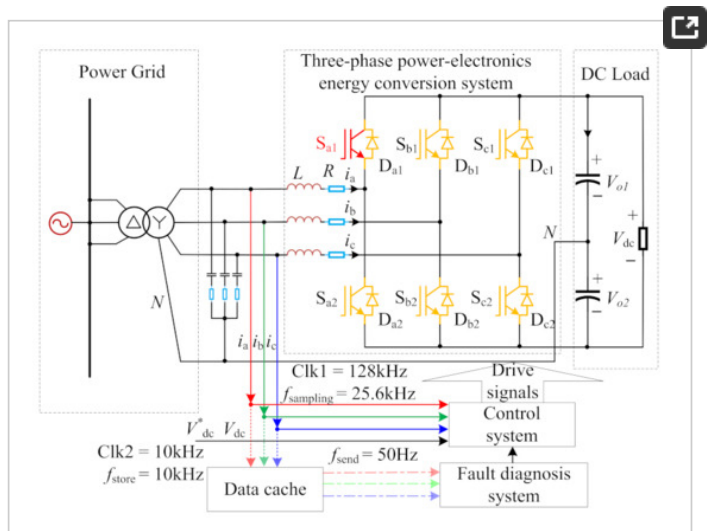

**Figure 14.** Fault diagnosis schematic for power electronic energy conversion systems.

*3.3. Monitoring for Submarine Cables*

Submarine cables are key components of offshore wind power transmission and play an important role in the development of offshore wind power [178]. The construction of offshore wind power projects inevitably involves a large number of submarine cables. As a concealed project, submarine cables are limited by the way the cables are installed and the uneven environmental temperature. With the increase of marine development activities, mechanical damage to submarine cables can also be caused by aquaculture, fishing nets, anchors and so on. Sea water erosion and other factors can easily cause poor water resistance performance and insulation aging of submarine cables. Once the submarine cables are damaged and stop operation, huge economic losses will result. Therefore, in order to ensure the safe operation of submarine cables, it is necessary to monitor the operation status of submarine cables in real time.

In order to ensure their safe operation, many scholars have studied the online monitoring of submarine cables [179]. Zhu et al. [180] proposed an online monitoring method for submarine oil-filled cables in the Hainan Interconnection project in which the current of each phase cable was selected as the measured signal. He et al. [181] proposed a dual terminal voltage video synchronization method to monitor submarine cables in the Zhoushan 500 kV interconnection project. In [180,181], the monitoring systems should also be tested with other long-distance submarine cables, and the real-time performance of the monitoring systems should be considered in the future. Chen et al. [182] proposed a Brillouin optical time domain analysis-based method in which the optical cable was adopted to monitor the temperature of submarine cables to ensure the stability of the system, but more actual operation data should be considered to verify the method. Lux et al. [183] proposed a depth of burial of submarine power cable formations monitoring method in which distributed temperature sensing, electric load data, and thermal models were used as the detection signal, but the influence of ambient temperature should be further studied in the future. Masoudi et al. [184] proposed a submarine cable condition monitoring method in which a distributed optical fiber vibration sensor was used to monitor the location and strain level of each point on the cable. Fouda et al. [185] proposed a time–frequency domain characteristic and SVM classifier-based method for submarine cables in which the vibration signals of optical fiber were used to detect malicious attacks. Xu et al. [186] proposed a method for monitoring submarine cables based on the temperature increase in optical fibers and developed an online monitoring system based on a BOTDR-based submarine cable online monitoring system. In [184,185,186], the interference of the harsh marine environment in the optical fiber signals should be considered in future research so as to improve the practical application value of the method. Zhao et al. [187] proposed a monitoring system based on BOTDR for 110 kV submarine cables in which the temperature/strain information was used to locate the faults, but the distributed temperature and train simultaneous measurement technology should be improved to make the method more practical in the future.

*3.4. Monitoring for Other Equipment*

In addition to offshore wind turbines, power electronic converters, and submarine cables, some scholars have studied offshore booster stations, sensors, uninterruptible power supply (UPS), offshore wind power structures, and so on [188,189].

The offshore booster station (as shown in **Figure 15**) is mainly used for the arrangement of the electrical system, safety system, auxiliary system, and other equipment, which can collect power from the offshore wind farm and then output it from the offshore wind farm after boosting. The marine environment of the offshore booster station requires the prevention of salt fog, damp and heat, and biological mold. In some places, it also requires resistance to strong typhoons and strong waves as well as the capacity to deal with the problem of high ultraviolet radiation. Yang et al. [190] proposed a corresponding fire protection scheme for offshore booster stations, on-land central control centers, and offshore wind turbines of the offshore wind farms; more and more comprehensive fire prevention schemes for equipment should also be considered to avoid immeasurable losses caused by omissions in the future.

**Figure 15.** Offshore booster station.

The UPS in offshore wind farms is mainly used in the control system, data acquisition, monitoring system, communication system, video monitoring system, fire alarm system, and so on. **Figure 16** shows the UPS monitoring system developed by Shanghai Dpin Electronic Technology Co., Ltd., Shanghai, China.

**Figure 16.** UPS (Uninterruptible Power Supply) monitoring system.

## 4. Operation and Maintenance of Smart Offshore Wind Farms

Compared with onshore wind farms, the environment of offshore wind farms is more complex as the influence of wind, wave, even extreme ice, typhoon, earthquake, and other load excitation on the equipment is more complex. **Figure 17** shows the operation and maintenance cost of offshore wind power. Generally, offshore wind farms are far away from land, the cost of operation and maintenance is higher than that of onshore wind farms, the management staff of the wind farms cannot evaluate the structure regularly, and the response time for the accident is far longer than that for onshore wind farms [191,192]. Therefore, it is of great importance to establish a reasonable operation and maintenance management scheme for the stable development of offshore wind farms.

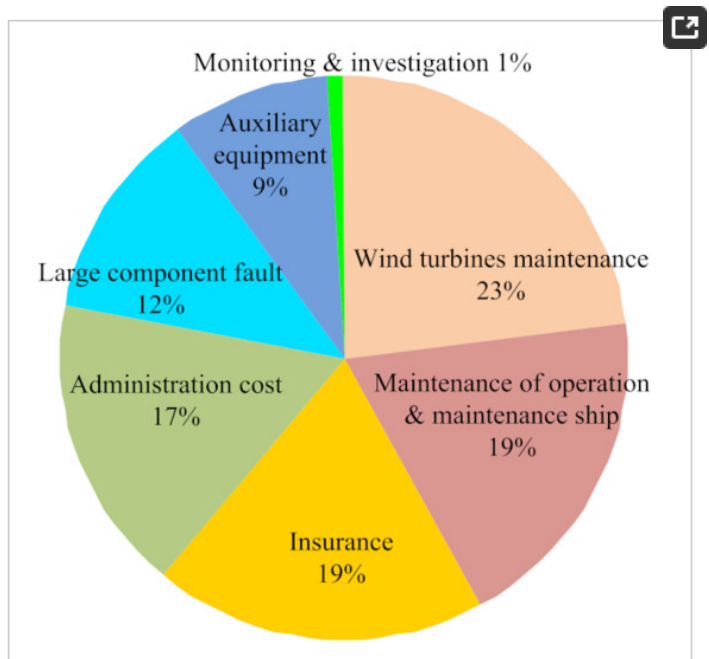

**Figure 17.** Operation and maintenance cost of offshore wind power.

*4.1. Operation and Maintenance Platform of Smart Offshore Wind Farms*

With the rapid development of global offshore wind power, the operation and maintenance demands of offshore wind farms also increase. The survey data, monitoring data, environmental parameters, and other different types of massive wind power data are constantly accumulating, which provide more reliable data for the construction, operation, and maintenance of offshore wind farms. The operation and maintenance management system, ships, robots, and big data platforms provide the basic guarantee for the stable and sustainable development of offshore wind power [193,194,195,196].

Artificial intelligence, big data, cloud computing, and several digital technologies play a very important role in the intelligent operation and maintenance platform of smart offshore wind farms. Lin et al. [193] proposed a deep learning neural network-based offshore wind power forecasting method in which data from the SCADA systems were adopted to construct the forecasting system so as to improve the quality of operation and maintenance. Yin et al. [194] proposed a deep neural learning (DNL)-based model predictive control (MPC) method (a hybrid CNN-LSTM model) in which the CNN-LSTM model was used to predict wind speed, wind turbine power, and other parameters. In [193,194], future research should consider feature extraction methods to eliminate redundant features. Wu et al. [195] proposed an AI technique-based method to optimize the arrangement of wind turbines in which the genetic algorithm (GA) and ant colony system algorithm were adopted to optimize the layout and line connection topology. Japar et al. [197] adopted five different machine learning methods (Support Vector Regression—SVR, linear regression, linear regression with feature engineering, ANN, and nonlinear regression) to estimate the power losses due to waves in large wind farms. In [195,197], the more practical operation factors (such as climate, environment, and other factors) of offshore wind farms should be considered in the future. Helsen et al. [198] adopted the big data approach to analyze the sensor data of different machines and the maintenance data, and the machine learning on SCADA data and pattern recognition methods were used to monitor offshore wind turbines to guarantee stable electricity production. However, future research should consider more data from other wind farms to develop a scalable and easy to promote platform system. Anaya-Lara et al. [199] adopted the SCADA systems to communicate with the operator, manufacturer, and maintenance crew as well as to remote control, regulate, and monitor modern wind farms. Since the faults of the network or sensors in offshore wind farms were due to harsh weather conditions, the SCADA data were often missing; thus, Sun et al. [200] proposed a learning framework to impute two missing-data conditions. Lin et al. [201] proposed an isolate forest (IF) and deep learning neural network-based method to reduce the impact of abnormal SCADA data. In [199,200,201], the problems of data encryption and abnormal data processing should also be deeply studied in the future, which are very important for the safe operation of offshore wind farms. As shown in **Figure 18**, the intelligent dispatching management system of offshore wind farms can integrate wind turbine monitoring, booster station monitoring, wind power prediction, ship scheduling, information management, and various equipment monitoring into a unified information platform, which can realize the integrated monitoring of offshore wind farms, evaluate the operation of offshore wind farms, provide a health warning, and greatly facilitate operation and maintenance.

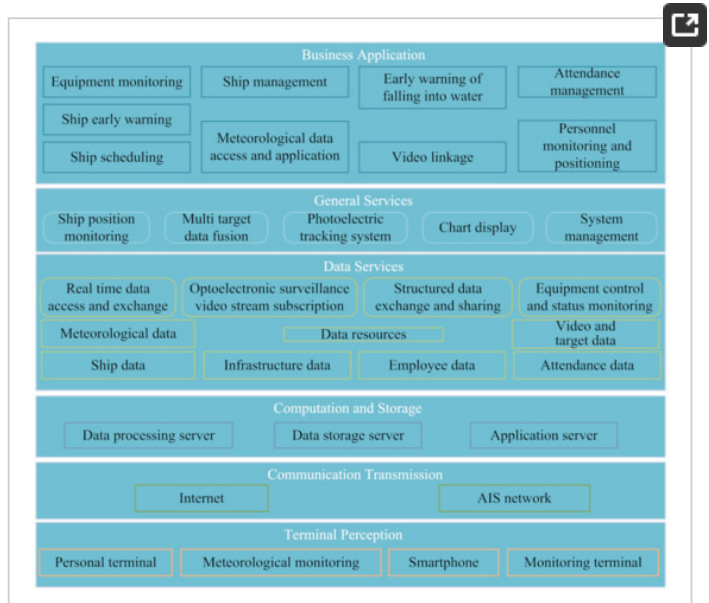

**Figure 18.** Intelligent dispatching management system of offshore wind farms.

At present, there are two main trends in the development of offshore wind farms. Wind farms are increasingly farther from the coast, require greater power generation, experience worse sea conditions, which bring more difficulties to their maintenance. The existing maintenance tasks for offshore wind turbines mainly include regular maintenance (inspection, cleaning, etc.), fault repair, equipment spare part management, etc. Therefore, wind power operation and maintenance ships, helicopters, and so on are essential for the daily maintenance of offshore wind farms (as shown in **Figure 19**), where the ship type directly affects their safety, rapidity, seakeeping, and maneuverability [202,203]. During the operation and maintenance of offshore wind farms, the transportation system can provide accommodation to the crew and technicians and can load, transport, and assemble the fault turbine components. Gundegjerde et al. [204] proposed a three-stage stochastic programming (SP) model to determine the ship fleet size and mix, and then to execute maintenance tasks in offshore wind farms. Stålhane et al. [205] proposed a two-stage SP model to determine which ships to charter and how to support maintenance tasks according to weather conditions and fault times. In [204,205], the cooperation of multiple ships and the optimization of the operation and maintenance path can also be considered in the future. In addition, unmanned intelligent equipment (such as unmanned boats and UAVs) has been developed rapidly, which provides a new choice for the operation and maintenance of smart offshore wind farms, and which has also been the development direction of offshore wind power operation and maintenance.

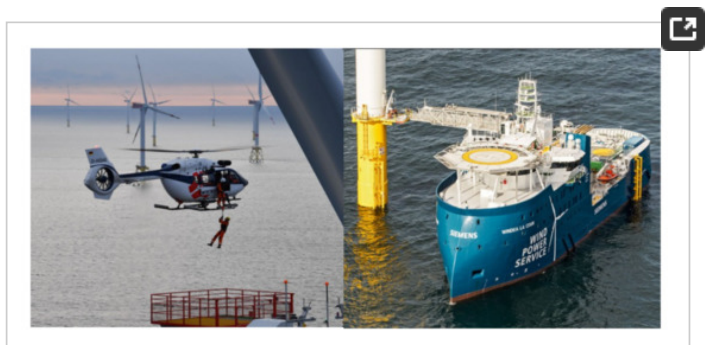

**Figure 19.** Transportation for the operation and maintenance of smart offshore wind farms.

*4.2. Operation and Maintenance Strategy for Smart Offshore Wind Farms*

In order to reduce the cost of operation and maintenance and improve the availability of offshore wind farms, it is necessary to scientifically and reasonably plan the operation and maintenance work for offshore wind farms so as to improve the quality and efficiency of operation and maintenance as well as reduce the attendance times and the cost of operation and maintenance [206].

Compared with onshore wind farms, the operation and maintenance of offshore wind farms are more affected by the environment and climate, and the operation and maintenance efficiency are lower. The operation and maintenance of offshore wind farms need to meet certain marine meteorological conditions. For example, when the wind speed is too fast or the waves are too high, operation and maintenance tasks cannot be completed. Martini et al. [207] analyzed the accessibility, approachability, weather window, and waiting time of offshore wind farms in the North Sea and subsequently made reasonable arrangements for their operation and maintenance; future research can also consider extending the research methods to other offshore wind farms so as to better optimize the methods. Lazakis et al. [208] analyzed the main maintenance influential factors of offshore wind farms (as shown in **Figure 20**) and proposed a heuristic optimization technique-based route planning and scheduling optimization framework to reduce the daily operation and maintenance costs, for which climate data, fault information, crew pick-up and drop-off tasks, wind farm attributes, and cost-related specifics were considered. Their research can also be optimized and adjusted according to the type of operation and maintenance personnel. Guo et al. [209] proposed an anti-typhoon control strategy (as shown in **Figure 21**), and the particle swarm optimization (PSO) and GA optimization algorithms were adopted to optimize the control strategy, which can improve the service life of wind turbines. Liu et al. [210] adopted a full-set three-dimensional meteorology simulation technique to simulate artificial typhoon wind fields, which can help with the design of typhoon-resistant schemes for offshore wind farms. In [209,210], future research should also consider more factors (such as the wind force and destructive force of typhoons) in the actual area to adjust the simulation and so as to make the method more practical. Ma et al. [211] selected a three-hour representative truncated typhoon wind speed data, and the blade element momentum (BEM) theory was adopted to study the effects of the NREL (National Renewable Energy Laboratory) 5 MW wind turbine control system and the floating platform on floating offshore wind turbine system; however, the robust control strategy for the floating offshore wind turbine systems still needs to be further enhanced when facing typhoon weather. Besnard et al. [212] proposed a cost-based optimization and selection model in which the number of technicians, transfer ships, helicopters, and the transportation strategy were taken into account. Wang et al. [213] proposed an ordered curtailment strategy for offshore wind farms based on the impact of typhoons, which can reduce the adverse effects of typhoons and reduce the operation costs. In [212,213], future research can consider extracting a historical record of an offshore wind farm's successful experience in order to optimize the model and strategy.

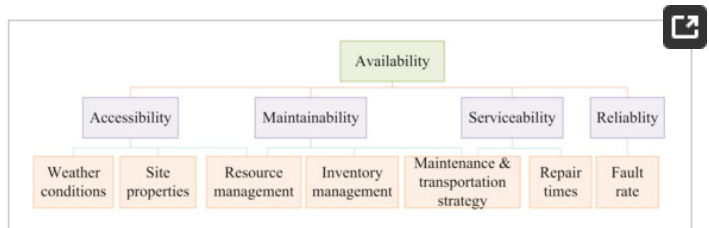

**Figure 20.** Main influential factors in the maintenance of offshore wind farms.

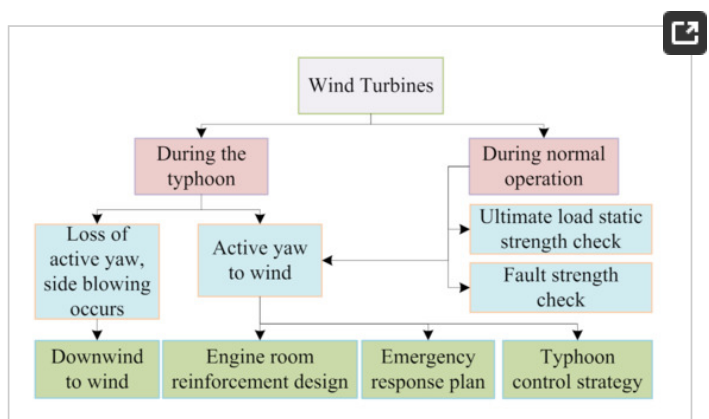

**Figure 21.** Anti-typhoon strategy for offshore wind farms.

The operation and maintenance strategies for offshore wind farms mainly include preventive maintenance and post repair; preventive maintenance mainly includes regular maintenance and status a sound opportunistic maintenance strategy to reduce the costs of operation and maintenance in which three types of maintenance opportunities (the age-based opportunity and the opportunities created by incident and degradation faults) were integrated to operate and maintain offshore wind farms. However, the study should also consider the actual operating equipment parameters and historical data so as to improve the operation and maintenance methods. Zhang et al. [216] developed a two-stage adaptive robust model to optimize daily maintenance tasks and production tasks; the column-and-constraint generation (C&CG) algorithm was used to decompose similar two-stage problems to a master problem and a sub-problem. Different transaction models and decision scenarios can be taken into account to optimize the maintenance method in the future. Kang et al. [217] introduced an opportunistic offshore wind farm maintenance policy with the consideration of the weather window effect and imperfect maintenance. Preventive maintenance was carried out for other devices, and some devices failed or reached the critical degradation states. In order to reduce loss from accidental faults and the maintenance costs, future research can consider predicting equipment lifetime by maintaining the equipment in advance. Yeter et al. [218] proposed a risk-based inspection and maintenance planning for offshore wind farms in which different inspection policies were studied, and the most cost-effective inspection and maintenance policy was selected; however, some actual cost components should be taken into account to better optimize the method in the future. As shown in **Figure 22**, Dalgic et al. [219] proposed a comprehensive operation and maintenance strategy to optimize the operation and maintenance costs, operation and maintenance tasks, transportation systems, revenue loss, and power production. Considering that the wind turbine systems are usually located in icy, cold, or remote offshore areas, and that the equipment ages due to long-term wear, corrosion, erosion, fatigue, and other factors, Shafiee [220] proposed an optimal age-based group maintenance strategy for offshore wind farms so as to reduce the operation and maintenance costs of offshore wind power, especially the high transportation and logistics costs. Sørensen [221] proposed a risk-based life cycle method to optimize the operation and maintenance plan in which the pre-posterior Bayesian decision theory was adopted for monitoring before the faults occur and to reduce the costs related to the monitoring, repair, maintenance, and so on. In [219,220,221], the aging and fault relationship between different components can be considered, and relevant information can be used for preventive operation and maintenance. Martin et al. [222] proposed a sensitivity analysis method to find the important factors related to operation and maintenance costs and availability; they found that the minor and major repair costs, operation duration, and the length of maintenance task were the important factors affecting the total operation and maintenance costs of offshore wind farms. Ahsan et al. [223] adopted the stakeholder analysis method to manage and coordinate with the various stakeholders related to the operation and maintenance in offshore wind farms; meanwhile, co-operation was adopted to improve the operation and maintenance efficiency and to reduce operation and maintenance costs. In [222,223], repair and maintenance can be considered at the same time so as to effectively reduce the frequency of offshore attendances and reduce operation and maintenance costs.

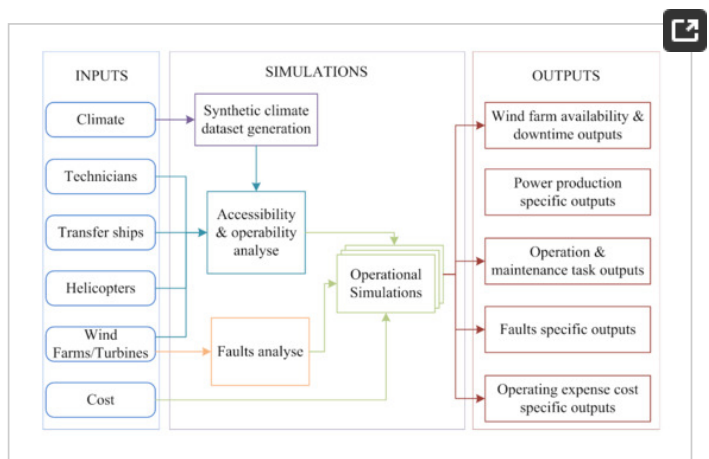

**Figure 22.** Operation and maintenance strategy.

*4.3. Safety and Management of Offshore Wind Farm Personnel*

The harsh environment makes operation and maintenance more difficult in offshore wind farms, but also brings great challenges to the operation and maintenance personnel. Offshore wind power maintenance personnel are usually scattered across different wind turbines or ships, and there are some potential risks such as falling from height, drowning, asphyxiation, poisoning in semi-enclosed spaces, electric shock, and so personnel before taking posts, but attention should also be given to the state of operation and maintenance personnel during operation and maintenance, and human errors should be avoided as much as possible in the operation and maintenance process of offshore wind farms [224].

In order to improve the rescue efficiency and reduce loss due to marine accidents, many scholars have studied the search scope, rescue methods, etc. As shown in **Figure 23**, it is necessary to consider the search areas, resource limitations, and search objects when designing and optimizing the search and rescue (SAR) activities. Xiong et al. [225] proposed a three-stage intelligent decision method to optimize the SAR plan in a maritime emergency, which can speed up SAR activities and reduce the loss of life. Atkinson [226] suggested strengthening the management of all kinds of ships (including the maximum number of passengers, working conditions, etc.), and meanwhile, it should cooperate with other regulatory agencies and industries to formulate unified standards and establish a complete offshore wind farm operation and maintenance scheme. In [225,226], more maritime emergencies should be considered in the future research. Zhou et al. [227] proposed a method for evaluating maritime search and rescue capability, and the response time of rescue ships was measured by the geographic information system (GIS)-based response time model; however, the response time of the SAR system must be deeply studied in the future, especially in extreme weather conditions. Deacon et al. [228] proposed a method based on major incident investigation and expert judgment techniques to evaluate the risks of human error in offshore emergency situations, which can reduce the rescue fault rate caused by human error. Nevertheless, more effective expert experience should be taken into account in the future. Skogdalen et al. [229] proposed some measures for the improvement of the evacuation, escape, and rescue operations when faced with offshore accidents, which can reduce the unnecessary losses caused by human errors. Liu et al. [230] proposed a helicopter-based maritime search and rescue method, which can better realize low-altitude search, hovering rescue, and to get people out of danger faster. In [229,230], when carrying out a rescue operation at sea, the state of the rescued object, weather conditions, and feasible means of transportation for rescue should be considered before making a comprehensive analysis and formulating a more reasonable rescue strategy.

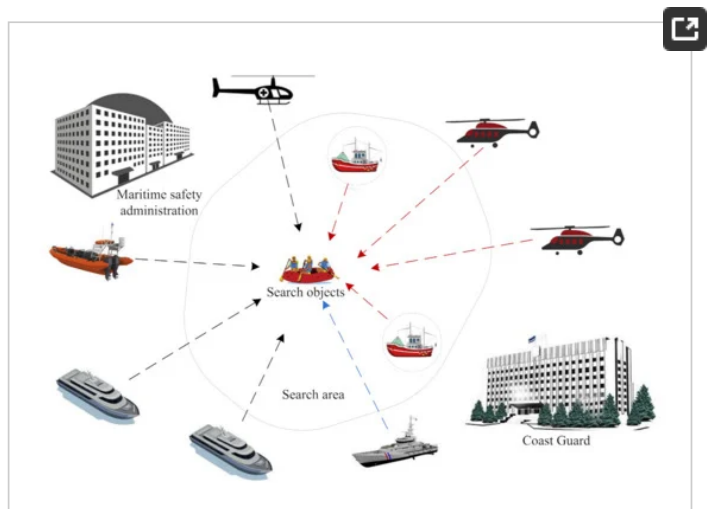

**Figure 23.** Search and rescue in a maritime emergency.

The smart dispatching system, the offshore wind power radar multi-source detecting and tracking system, the boundary warning system, and the operation supervision system of offshore wind turbine platforms were used to ensure the safety of ships, operation and maintenance personnel, wind turbines, and submarine cables. As shown in **Figure 24**, Liu et al. [231] proposed a method for monitoring the working state of operation and maintenance personnel, which can provide the guidance maintenance strategies according to the physiological signals of operation and maintenance personnel and reduce human errors; however, age, gender, and other factors should also be considered when dividing the tensions of operation and maintenance personnel. Due to the shortage of offshore wind power operation and maintenance personnel, the operation and maintenance capacity is insufficient. Additionally, there are many offshore operation types that include the basic inspection of offshore wind turbines and offshore booster stations, and other equipment need high professional operation and maintenance ability. Therefore, the comprehensive ability and technical level of operation and maintenance personnel should be improved. The offshore wind power industry has a strong particularity, especially as offshore communication conditions are relatively poor, and there are some blind areas in communication and exchange which increase the security risks of the operation and maintenance personnel. Therefore, in the process of employing the operation and maintenance staff, it is necessary to ensure that they have more professional skills; the safety training for operation and maintenance staff should also be carried out to improve their awareness of safety and their ability to investigate potential danger.

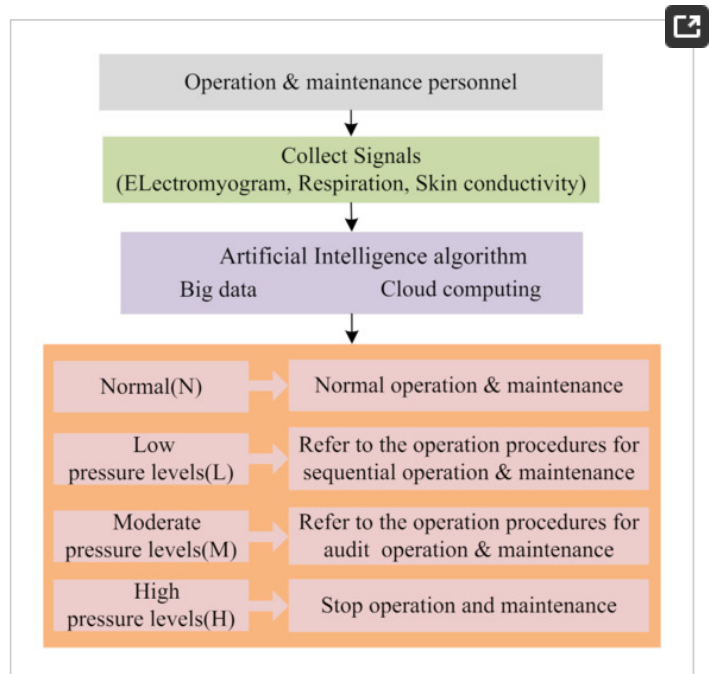

**Figure 24.** Operation and maintenance based on the pressure of operation and maintenance personnel.

## 5. Conclusions and Prospects

This paper summarized the research on the monitoring, operation, and maintenance of smart offshore wind farms. The environmental monitoring technologies, some advanced equipment and technologies, some power equipment monitoring methods, and the operation and maintenance strategies for smart offshore wind farms were discussed in detail. In order to improve the stability of offshore wind farms, to improve the quality and efficiency of operation and maintenance, and to increase the revenue of offshore wind farms, this paper puts forward the following research points and trends:

1. During the construction of offshore wind farms, it is necessary to monitor the marine environment and marine organisms for a long time, and to try to avoid or reduce the impact on the habitats and migration routes of birds, fish, and other marine organisms. At the same time, the integration of offshore wind farms and marine ranches can be considered to realize the efficient output of clean energy and safe aquatic products, which will be an important industrial mode and future development direction.

2. Due to the high cost of operation and maintenance helicopters and ships, the advanced data analysis platform, model display platform, and visualization platform should be considered, which can make full use of the accumulated operation data to predict and analyze the state of the offshore wind power equipment, so as to scientifically carry out the operation and maintenance of offshore wind farms, to fully realize predictive maintenance and intelligent maintenance for offshore wind power equipment, to optimize the frequency of operation and maintenance, and to reduce the operation and maintenance cost.

3. In the power equipment intelligent monitoring field, the current intelligent monitoring method relies too much on data samples. In addition, the domain knowledge-driven method can be employed, which can reduce the dependence on data samples. In particular, some expert experience and knowledge can be used for feature extraction, which can effectively reduce the dependence on data samples of different operation conditions.

4. In a long-distance sea voyage, the special operation and maintenance ship is likely to be affected by the weather and sea conditions. For example, when the operation and maintenance ship sets out, the sea state is still calm, but it has to turn back due to the sudden change in weather halfway to the operation site, prediction capabilities for regional climate and weather at the offshore wind farms and to provide real-time weather information for the reasonable planning, operation, and maintenance of offshore wind farms so as to reduce unnecessary operation and maintenance times and costs.

**Author Contributions**

Conceptualization, L.K. and Y.L.; methodology, F.Z.; software, L.K.; validation, L.K., Y.L. and F.Z.; formal analysis, W.K.; investigation, L.K.; resources, F.Z.; data curation, X.G., W.K. and Q.Y.; writing—original draft preparation, L.K., Y.L. and F.Z.; writing—review and editing, Y.H., X.G., W.K. and Q.Y.; visualization, X.G. and Q.Y.; supervision, Y.H.; project administration, F.Z., Q.Y. and W.K.; funding acquisition, Q.Y. and F.Z. All authors have read and agreed to the published version of the manuscript.

**Funding**

This research is partly supported by the science and technology projects of the Jilin Province Department of Education (JJKH20191262KJ and JJKH20191258KJ).

**Institutional Review Board Statement**

Not applicable.

**Informed Consent Statement**

Not applicable.

**Data Availability Statement**

Not applicable.

**Acknowledgments**

We would like to thank the Institute of Oceanographic Instrumentation, Shandong Academy of Sciences; Alphaer (Shenzhen) Technology Co., Ltd.; and Shanghai Dpin Electronic Technology Co., Ltd. for their support and help, and for providing us with the rights to use the equipment and pictures, for which there is no copyright.

**Conflicts of Interest**

The authors declare no conflict of interest.